%% file: template.tex
\documentclass{article}

\usepackage{arxiv}

\usepackage[utf8]{inputenc} 
\usepackage[T1]{fontenc}    
\usepackage{hyperref}       
\usepackage{url}            
\usepackage{booktabs}       
\usepackage{amsfonts}       
\usepackage{nicefrac}       
\usepackage{microtype}      
\usepackage{lipsum}
\usepackage{graphicx}
\graphicspath{ {./images/} }

\input{notation.tex}

\usepackage{algorithm}
\usepackage{algorithmic}
\usepackage{amsmath}
\usepackage{dsfont}

\title{ABC: Aggregation before Communication, a Communication Reduction Framework for Distributed Graph Neural Network Training and Effective Partition}

\author{
Junwei Su\\
Computer Science Department, The Unversity of Hong Kong\\
junweisu@connect.hku.hk
}

\begin{document}
\maketitle

\input{tex_content/paper/abstract.tex}
\input{tex_content/paper/introduction.tex}
\input{tex_content/paper/background.tex}
\input{tex_content/paper/communication.tex}

\input{tex_content/paper/partition.tex}
\input{tex_content/paper/online.tex}
\input{tex_content/paper/conclusion.tex}

\bibliographystyle{unsrt}  
\bibliography{references}

\end{document}

%% file: notation.tex
\newtheorem{definition}{Definition}

\newtheorem{theorem}{Theorem}

\newcommand{\graph}[0]{\mathds{G}}



%% file: tex_content/paper/abstract.tex
\begin{abstract}
    Graph Neural Networks(GNNs) are a family of neural models tailored for graph-structure data and have shown superior performance in learning representations for graph-structured data. However, training GNNs on large graphs remains challenging and a promising direction is distributed GNN training, which is to partition the input graph and distribute the workload across multiple machines. The key bottleneck of the existing distributed GNNs training framework is the across-machine communication induced by the dependency on the graph data and aggregation operator of GNNs. In this paper, we study the communication complexity during distributed GNNs training and propose a simple lossless communication reduction method, termed the Aggregation before Communication (ABC) method. ABC method exploits the permutation-invariant property of the GNNs layer and leads to a paradigm where vertex-cut is proved to admit a superior communication performance than the currently popular paradigm (edge-cut). In addition, we show that the new partition paradigm is particularly ideal in the case of dynamic graphs where it is infeasible to control the edge placement due to the unknown stochastic of the graph-changing process.
\end{abstract}

%% file: tex_content/paper/introduction.tex
\section{INTRODUCTION}\label{sec:introudction}

Graph Neural Networks (GNNs) is a neural model tailored for graph-structure data and have become one of the fastest growing subareas in deep learning in the recent past~\cite{grl}. The key differentiation between GNNs and other neural models is the usage of the graph structure to facilitate information propagation. Each layer of GNNs composite of a two-step aggregation-update procedure. Given a target vertex, GNNs layers first {\it aggregate} features or intermediate representations from its neighbour vertex set. Then, the representation of the targeted vertex is {\it updated} with the aggregated result via a multi-layer perceptron. Such an aggregation-update procedure allows GNNs to combine and learn information from both the graph structure and feature.


To facilitate the aggregation-update procedure, a k-layer GNNs need to create a k-hop computation for each node based on its neighbourhood. Because of this, despite the promising performance, training GNNs at scale remain challenging, as the amount of computation and memory resources needed for training a real-world web-scale graph is prohibitive. Distributed GNN training has emerged as an important and promising direction to advance the practicality of GNNs by allowing them to be trained on a large graph. The standard procedure of distributed GNN training first partitions a giant graph into multiple small subgraphs, each of which is able to fit into a single machine (also referred to as a worker), and then each machine trains these partitioned subgraphs locally together with indispensable communication across partitions induced by the dependency of graph-structured data and the aggregation-update nature of GNNs. In addition, GNNs models are extremely light-weighted, rendering the main bottleneck in the training process mostly induced by the communication of node features.

In this work, we propose a novel communication reduction mechanism for distributed GNNs training, termed Aggregation Before Communication (ABC), which is based on the permutation invariant property of GNN layers and can reduce communication costs across devices without the loss of any information. The usage of ABC also leads to a new graph partition paradigm targeting communication reduction with a theoretical guarantee. Our main contribution can be summarized as follows:

\begin{enumerate}
    \item Based on the permutation-invariant property, we theoretically show that the aggregation operation can be broken down into sub-recurrent processes with inputs in arbitrary order. 
    \item Based on the insight above, we proposed a new communication protocol across devices, termed ABC,  which can reduce the communication cost without any loss of information and negligible extra cost. The proposed method is compatible with existing communication reduction methods such as quantization. We prove a guarantee of performance improvement based on the structural property of the partitioned graph.
    \item The performance analysis of the proposed communication method leads to a new graph partition scheme tailored for communication reduction. The new induced partition scheme is based on vertex-cut, and it is proved to have an improvement of at least two-fold compared to the currently popular edge-cut partition scheme. 
    \item We show that the new communication and partition scheme discussed in this paper is particularly suitable for the case of dynamic graphs where it is infeasible to control the edge placement for future data due to a lack of knowledge in the future. Our scheme provides a simple objective for the online partition.
\end{enumerate}

%% file: tex_content/paper/background.tex
\section{BACKGROUND}\label{sec:background}

\subsection{Graph Neural Network}
GNNs learn the embedding of each node in a graph via a two-step aggregation-update process (neighbour aggregation and then node update) for each layer, which can be mathematically described as
\begin{equation}
    z_{v}^{(l)} = \text{AGGREGATE}^{(l)}(\{h_u^{(l-1)} |u \in N(v)\})
\end{equation}
\begin{equation}
    h_v^{(l)} = \text{UPDATE}^{(l)}(z_{v}^{(l)}, h_v^{(l-1}))
\end{equation}

where UPDATE and AGGREGATE are arbitrary differentiable functions (i.e., neural networks) and $z_v^{(l)}$ is the “message” that is aggregated from u’s graph neighbourhood $N(u)$. The essential difference between these representative GNNS (as well as other existing ones) is the different formulations of the aggregate-update procedure. However, because of the orderless nature of graph-structured data (e.g., there should not be any ordering of the neighbourhood for a given vertex), GNNs layers need to satisfy either permutation-invariant or permutation-equivariant property~\cite{grl}. Permutation-invariant property means that the output of a function should be independent of the order of the inputs. Permutation-equivariant property means that the output of a function should be permutated according to the permutation of the inputs. Most (almost all) existing GNNs, including representative GCN~\cite{gcn}, GraphSage~\cite{sage}, and GAT~\cite{gat}, satisfy the permutation-invariant property.  The permutation-invariant property of GNN layers is the basis for our proposed communication reduction method and should be compatible with future GNNs, as they are likely to obey the permutation-invariant property as well.

\subsection{Distributed GNN Training}
Existing frameworks for training GNNs, such as the Deep Graph Library (DGL)~\cite{distDGL,dgl}, support distributed training by combining distributed graph processing techniques with DNN techniques, as shown in Fig.~
The input graph and features are partitioned across machines in the cluster. Given a batch size (1), the computation graph for each node, commonly referred to as a training sample, in the batch is generated by pulling the k-hop neighbourhood of each node along with the associated features (2). This requires communication with other machines in the cluster. Once the computation graphs are constructed, standard DNN techniques such as data parallelism are used to execute the computation—minibatches are created and copied to the GPU memory (4). Then the model computation is triggered (5).

The main bottleneck in the above procedure is (2) which involves cross-machines communication of the required node features for computation~\cite{distDGL,p3,pipegcn}. There are existing works tackling this problem with various approaches. ~\cite{vqgnn} adopt the quantization strategy to reduce the size of communication at the cost of loss of information. ~\cite{distDGL} duplicate the extra one-hop information for each partition to reduce the communication requirement. However, this method comes at a huge cost of extra memory storage and become ineffective when the layer of GNN is larger than one. ~\cite{p3,pipegcn} adopt a pipeline framework, similar to the ones used in DNN distributed training, to overlap the communication and computation stages. The frameworks and mechanism used in ~\cite{p3,pipegcn} are complicated as it requires to changing the underlying learning dynamic (optimization algorithm). In addition, the communication cost is much larger than the computation cost in distributed GNN training, as GNN models are very light-weighted. This means that the effect of reduction resulting from overlapping communication and computation is likely to be limited. 

%% file: tex_content/paper/communication.tex
\section{ABC COMMUNICATION PROTOCOL}\label{sec:proposed}

The AGGREGATE operator in GNN layers can be viewed as a function $\mathbf{A}$ that maps a list of vectors into a single representative vector. More formally, we can define the AGGREGATE operator as follows.

\begin{definition}\label{def:aggregator}
    Let $\rho (\mathbb{R}^k)$ denote the powerset of all possible vectors in $\mathbb{R}^k$. Then the AGGREGATE operator $\mathbf{A}$ can be defined as a function,
    \begin{equation}
        \mathbf{A}: \rho (\mathbb{R}^k) \mapsto \mathbb{R}^k
    \end{equation}
\end{definition}

We have the following result for a permutation-invariant AGGREGATE operator.

\begin{theorem}\label{thm:invariant}
    For all permutation-invariant AGGREGATE operator $\mathbf{A}$ and a finite non-intersecting partitions $\{B_i\}$, there exist functions $f$ and $g$ such that 
    \begin{equation}
        \mathbf{A}(\bigcup_i B_i) = g(\bigcup_i f(B_i))
    \end{equation}
\end{theorem}

The proof of Theorem~\ref{thm:invariant} relies on Kolmogorov–Arnold representation theorem~\cite{kolmogorov} and the permutation-invariant property. Theorem~\ref{thm:invariant} implies that so long as the AGGREGATE operator is permutation-invariant, we can compute the end result by arbitrarily dividing the input into a series of subsets and perform local aggregations ($f$) followed by a global aggregation ($g$).

{\bf EXAMPLE 1:} let's consider a one-layer GNN with representative AGGREGATE operator sum $\sum$ and a vertex $u$ with three neighboring vertice, $v,w,z$. Furthermore, let $f$
 and $g$ be the same operator sum $\sum$ in this case. The original aggregation for $u$ is $\sum \{x_v,x_w,x_z\}$ which is equal to $\sum \{ \sum\{x_v,x_w\},x_z\}$ or $\sum \{ x_v,\sum\{x_w,x_z\} \}$ or $\sum \{ \sum\{ x_v,x_z\},x_w \}$.

Obviously, this example can be generalized to arbitrary inputs. In addition, it is straightforward to construct $f$ and $g$ for the AGGREGATE operators in the GNN design space~\cite{gnn_ds}. As discussed earlier, other potential AGGREGATE operator (in the future) of GNN would also need to satisfy the permutation-invariant property and therefore, Theorem~\ref{thm:invariant} should still be applicable. Next, we introduce some notions to be used in the algorithms and theoretical analysis later.

\begin{definition}
    Given a vertex $v$ with immediate neighborh set $N(v) = N(v)^l \cup N(v)^c$, we define $N(v)^l \subset N(v)$ to be the set of local neighbour vertices that are in the same worker as $v$ and $N(v)^c \subset N(v)$ to be the set of cross neighbour vertices that are in a different worker from $v$.
\end{definition}

\begin{definition}\label{def:boundary_vertex}
    We call a vertex $v$ a {\it boundary vertex} if $N(v)^c \neq \emptyset$.
\end{definition}

Furthermore, we denote the complete set of boundary vertices in worker $w_i$ as $\mathcal{B}(w_i)$, i.e. 
$$\mathcal{B}(w_i) = \{u \in \graph_i | N(u)^c \neq \emptyset \}.$$ 

In this paper, we are interested in the number of messages as the communication complexity, and we denote that the communication complexity from worker $w_j$ to worker $w_i$ as $C(w_i,w_j)$,
$$C(w_i,w_j) = \mathcal{M}_{ij} = \text{number of messages } w_j \text{ send to } w_i .$$

In the standard procedure for distributed GNN training introduced in Sec.~\ref{sec:background}, the communication cost is induced by the vertices in the boundary set of each worker sending requests to the other workers asking for its cross neighbour vertices to complete the computation. The communication complexity $C_{s}(w_i,w_j)$ or the set of message $\mathcal{M}_{ij}^{s}$ that $w_i$ needed from worker $w_j$ under the standard procedure can be expressed as,

\begin{equation}\label{eq:edge_cut_message}
    C_{s}(w_i,w_j) = \mathcal{M}_{ij}^{s} = \{ x_u | u \in N(v)^c, v \in \mathcal{B}(w_i)\} 
\end{equation}

It is obvious that $C_{s}(w_i,w_j)$ is correlated and bounded the number of edge between worker $w_i$ and other workers. Therefore, it makes sense for the standard distributed GNN training to consider minimizing the edge-cut for partitioning the graph. Next, we present a better communication strategy (ABC) based on Theorem~\ref{thm:invariant}.  The idea of ABC communication strategy is straightforward, which is to decompose the AGGREGATE operator of GNN into local aggregation $f$ and global aggregation $g$, and to delegate the local aggregation $f$ to the worker that is sending the message. As the main bottleneck in distributed GNN training is the cross-worker transfer, doing so allow for a favorable trade-off between the communication complexity (number of message to be sent) and the computation complexity (performing extra aggregations). For clarity and simplicity, in the rest of the paper, we focus on the setting with a one-layer GNN and two workers $w_1, w_2$. Let $f$ and $g$ be the local and global aggregation function associated with the given GNN. The communication protocal of SEND and RECEIVE of ABC in this case are given in Algorithm~\ref{alg:abc_send} and Algorithm~\ref{alg:abc_receive}.

\begin{algorithm}
\caption{SEND: Protocol for Sending Messages for $w_j$}\label{alg:abc_send}

\textbf{RECEIVE:} a set of request $\mathcal{R} = \{v_1,v_2,...,v_m\}$ from $w_i$  \;

\textbf{SEND:} $\mathcal{M}_{i,j}$ to $w_i$ 
$$\mathcal{M}_i = \{f(N(u)^c) | u \in \mathcal{R}\}$$
\end{algorithm}

\begin{algorithm}
\caption{SEND: Protocol for Requesting Messages for $w_i$}\label{alg:abc_receive}

\textbf{SEND:} a set of request $\mathcal{R} = \{v_1,v_2,...,v_m\}$ to $w_j$

\textbf{RECEIVE:} a set of requested feature with aggregation $\mathcal{M}_i = \{f(N(u)^c) | u \in \mathcal{R}\}$  \;

\textbf{COMPUTATION:} perform the rest of computation with the local aggregation results and the global aggregation function $g$. 
\end{algorithm}

Next, we present a theoretical analysis on the communication complexity under the proposed communication protocol.

\begin{theorem}\label{thm:abc_communication_complexity}
    Under Algorithm~\ref{alg:abc_send}, the cross-worker communication complexity from $w_i$ to $w_j$ (in term of number of messages) is bounded by $|B(w_j)|$.
\end{theorem}

Suppose $E_c$ is the number of cross-worker edge between worker $w_1$ and $w_2$

\begin{theorem}\label{thm:improvement}
    Algorithm~\ref{alg:abc_send} to admit an improvement comparing to the standard approach given in Eq.\ref{eq:edge_cut_message} if there exists a worker $i$ such that 
    $|B(w_i)| < E_c.$
    Furthermore, the improvement is lower bounded by $E_c - |B(w_j)|$.
\end{theorem}

\subsection{Extending to k-layer GNN with m Workers}
In this subsection, we dicuss how the proposed communication algorithm and the derived result can be extended to the general casew with k-layer GNN and m workers. Because of the back-propagation, all the immediate representations are needed for computing the gradient. However, the communication protocol from Algorithm~\ref{alg:abc_send} and Algorithm~\ref{alg:abc_receive} can still be applied to optimize the communication for the features/messages needed for the last layer of GNN, which takes up the majority part of the communication. Extending the analysis to this general case is straight-forward and would be in future work.

%% file: tex_content/paper/partition.tex
\section{GRAPH PARTITION}\label{sec:partition}
An interesting perspective on the theoretical results above is that under the PAC communication method, the communication is decoupled from the edge cut between different machines. Instead, the boundary vertices become the key topological property for deciding cross-machine communication. Based on this insight, we propose a simple partition algorithm based on vertex-cut and theoretically analyze how the communication with the proposed partition algorithm performs against the current partition paradigm (edge-cut) for distributed GNNs training.

\begin{definition}\label{def:vertex_cut}
    Given a connected graph $\graph$, we say $\mathcal{V}_c \subset V(\graph)$ a vertex-cut if $\graph - \mathcal{S}$ contains more than one connected components.
\end{definition}

\begin{algorithm}
\caption{Vertex-Cut Based Graph Partition}\label{alg:partition}
\textbf{INPUT:} a connected graph $\graph$, and a vertex-cut $\mathcal{V}_c$

Randomly partition the vertices from $\mathcal{V}_c$ into two even partition $\mathcal{V}_1,\mathcal{V}_2$ (off by at most one vertex).

Let $\mathcal{E}_{\mathcal{V}_1,\mathcal{V}_2}$ denote the edges between $\mathcal{V}_1,\mathcal{V}_2$ and  $\mathcal{E}_{\mathcal{V}_1,\mathcal{V}_2}$ forms a edge-cut of graph $\graph$.

Partition the graph based on $\mathcal{E}_{\mathcal{V}_1,\mathcal{V}_2}$
\end{algorithm}

Given a graph $\graph$, let $\mathcal{V}_c^*$ denote the optimal vertex-cut and let $\mathcal{E}_c^*$ denote the optimal edge-cut. We have the following relation between $\mathcal{V}_c^*$ and $\mathcal{E}_c^*$.

\begin{theorem}\label{thm:vc_ec_rel}
    Given a connected graph $\graph$, let $\mathcal{V}_c^*$ denote the optimal vertex-cut and $\mathcal{E}_c^*$ denote the optimal edge-cut. We have
    \begin{equation}
            |\mathcal{V}_c^*| \leq |\mathcal{E}_c^*|.
    \end{equation}
\end{theorem}

\begin{theorem}\label{thm:partition_guarantee}
    Given a connected graph $\graph$ and its optimal vertex-cut $\mathcal{V}_c^*$, let $\graph_1$ and $\graph_2$ be two partitioned graph returned by Algorithm~\ref{alg:partition} and let $B^* = \max \{\mathcal{B}(w_1),\mathcal{B}(w_2)\}$. Then, we have
    \begin{equation}
        B^* \leq \frac{|\mathcal{V}_c^*|}{2}
    \end{equation}
\end{theorem}

\begin{theorem}\label{thm:communication_vertex_cut}
    Given a connected graph $\graph$ and its optimal vertex-cut $\mathcal{V}_c^*$, let $\graph_1$ and $\graph_2$ be two partitioned graph returned by Algorithm~\ref{alg:partition} and let $B^* = \max \{\mathcal{B}(w_1),\mathcal{B}(w_2)\}$. Then, under the proposed communication algorithm~\ref{alg:abc_send}, we have the 
    $$C(w_i,w_j) \leq \frac{|\mathcal{V}_c^*|}{2}, \forall i,j$$
\end{theorem}

\begin{theorem}\label{thm:improvement}
   Under communication Algorithm~\ref{alg:abc_send}, vertex-cut partition scheme with Algorithm~\ref{alg:partition} admits an improvement (in term of communication message number) to the edge-cut partition scheme of at least two-folds.
\end{theorem}

Theorem~\ref{thm:improvement} shows that under the proposed communcation protocol, the vertex-cut based partition algorithm (even with our naive partition strategy) should be favored over the the edge-cut based partition algorithms. In addition, the improvement of this scheme for communication is substantial, rendering that this should be ``de-facto'' for distributed GNN training framework.

%% file: tex_content/paper/online.tex
\section{EXTENSION TO ONLINE PARTITION}\label{sec:evaluation}
The analysis and dicussion above is based on the case of static graph, where the underlying graph structure does not change during the training process. However, the proposed scheme of communication and partition criteria is particularly desired for online partition algorithm of dynamic graph. 

In the problem of online partitioning of dynamic graph, graph data comes in a streaming manner, meaning that more and more graph data would come in as time proceed. The challenge in online partition algorithm for dynamic graph is that the optimal solution at time $t$ might become sub-optimal and even a bad partition in the future when more graph data comes in. For example, in the edge-cut scheme, the boundary vertices in different workers might have a different interaction intensity over time. If the interaction intensity become active between two workers in the coming time period, then the previous partition become inferior. To maintain a good performance with a edge-cut partition scheme might need to actively migrate data across different workers, which is very costly. On the other hand, under the vertex-cut partition scheme, the communication cost between different workers is solely bounded by the boundary vertex set. This gives a novel objective for online dynamic graph partition algorithm, which is to balance the minimize the boundary vertex set of each worker.

%% file: tex_content/paper/conclusion.tex
\section{CONCLUSION}~\label{sec:conclusion}
In this paper, we propose the ABC communication reduction method based on the permutation-invariant property of GNN layers. The ABC communication reduction method is compatible with all existing distributed GNN training frameworks as it only needs to add another processing layer before sending a communication message and it does not alter any subsequential computation result. We present a thorough theoretical analysis of the proposed method and show that it naturally leads to a vertex-cut based graph partition scheme. We show that the combination of the proposed communication method and its partition scheme is very desirable in the case of dynamic graphs. In this paper, we conduct the theoretical analysis in a simplified setting, a one-layer GNN with two training machines. It is actually easy to extend the theoretical results to the general case of k-layer GNN with m training machines. For future work, we consider conducting an empirical study to further validate the claims and results presented in this paper.